\documentclass[letterpaper, 10 pt, conference]{ieeetran}  
\IEEEoverridecommandlockouts
\usepackage{cite}
\usepackage{amsmath,amssymb,amsfonts}
\usepackage{algorithmic}
\usepackage{graphicx}
\usepackage{textcomp}
\usepackage{xcolor}

\usepackage{booktabs}
\usepackage{url}
\usepackage{hyperref}
\usepackage{xcolor}
\usepackage{cleveref}
\usepackage{upgreek,bm}

\usepackage{comment}
\usepackage[ruled,vlined]{algorithm2e}
\usepackage{caption}
\usepackage{subcaption}
\usepackage{multirow}

\usepackage{array}
\newcommand{\PreserveBackslash}[1]{\let\temp=\\#1\let\\=\temp}
\newcolumntype{C}[1]{>{\PreserveBackslash\centering}p{#1}}
\newcolumntype{R}[1]{>{\PreserveBackslash\raggedleft}p{#1}}
\newcolumntype{L}[1]{>{\PreserveBackslash\raggedright}p{#1}}

\newcommand{\bx}{\mathbf{x}}
\newcommand{\bz}{\mathbf{z}}

\newcommand{\bsig}{\boldsymbol{\sigma}}
\newcommand{\bmu}{\boldsymbol{\mu}}
\newcommand{\ba}{\bm{q}}
\newcommand{\bb}{\bm{e}}
\newcommand{\bo}{\mathbf{o}}
\newcommand{\bbb}{\bb_{\text{target}}}
\DeclareMathOperator*{\FK}{\operatorname{FK}}

\newcommand{\norm}[1]{\left\lVert#1\right\rVert}

\renewcommand{\baselinestretch}{0.865}

\def\BibTeX{{\rm B\kern-.05em{\sc i\kern-.025em b}\kern-.08em
    T\kern-.1667em\lower.7ex\hbox{E}\kern-.125emX}}

\begin{document}

\markboth{IEEE Robotics and Automation Letters. Preprint Version. Accepted February, 2022}
{Hung \MakeLowercase{\textit{et al.}}: Reaching Through Latent Space} 

\title{\LARGE \bf Reaching Through Latent Space: From Joint Statistics to Path Planning in Manipulation}

\author{Chia-Man Hung$^{1,2}$, Shaohong Zhong$^{1}$, Walter Goodwin$^{1,2}$, \\
Oiwi Parker Jones$^{1}$, Martin Engelcke$^{1}$, Ioannis Havoutis$^{2}$, Ingmar Posner$^{1}$
\thanks{$^{1}$Applied AI Lab (A2I), $^{2}$Dynamic Robot Systems (DRS)}%
\thanks{Oxford Robotics Institute (ORI), University of Oxford}
\thanks{Correspondence to: {\tt\footnotesize chiaman@robots.ox.ac.uk}}%
}

\maketitle
%

\begin{abstract}

We present a novel approach to path planning for robotic manipulators, in which paths are produced via iterative optimisation in the latent space of a generative model of robot poses. Constraints are incorporated through the use of constraint satisfaction classifiers operating on the same space. Optimisation leverages gradients through our learned models that provide a simple way to combine goal reaching objectives with constraint satisfaction, even in the presence of otherwise non-differentiable constraints. Our models are trained in a task-agnostic manner on randomly sampled robot poses. In baseline comparisons against a number of widely used planners, we achieve commensurate performance in terms of task success, planning time and path length, performing successful path planning with obstacle avoidance on a real 7-DoF robot arm. 

\end{abstract}



\section{INTRODUCTION}

\IEEEPARstart{P}{ath} planning is a cornerstone of robotics. For a robotic manipulator, this generally consists of producing a sequence of joint states the robot needs to follow in order to move from a start to a goal configuration. This requires that the poses along the sequence are kinematically feasible while at the same time avoiding unwanted contact either by the manipulator with itself or with potential objects in the robot's workspace. Due to its importance, path planning is a richly explored area in robotics (e.g. \cite{lavalle1998rapidly, Kavraki1996prob, karaman2011sampling, ratliff2009chomp, KalakrishnanSTOMP2011, ratliff2018riemannian}). However, traditional approaches are often marred by a number of issues. As the state-space dimensionality increases and constraints become more constrictive, the decreasing efficiency of traditional planning methods makes reactive behaviour computationally challenging. While existing sampling and optimisation-based approaches to the planning problem can find solutions, they scale super-linearly with a robot's degrees of freedom, and those that have optimality guarantees on resulting paths are guaranteed to achieve this only \textit{asymptotically}, after infinite time \cite{karaman2011sampling}. Increasing system and task complexity also requires consideration of multiple objectives (e.g. performing a certain task while adhering to pose constraints). Yet, enforcing constraints on the planned motion can be difficult. Traditional optimisation-based planners can struggle to incorporate constraints that cannot be expressed directly in joint space. Sampling-based planners, on the other hand, struggle to find solutions in scenarios where constraints render only a small volume of configuration space feasible or where narrow passages exist \cite{Berenson2009}.

The advent of deep learning has shown that learning-based approaches can offer some relief in overcoming robotic planning and control challenges. While a considerable body of work examines the direct learning of control policies, attempts have been made to apply deep learning to robotic path planning (e.g. \cite{levine2018learning}). Learnt heuristics and neural network collision detectors have been used as drop-in replacements to stages of traditional methods (e.g. \cite{Ichter2019, Qureshi2019, Qureshi2020}). A number of works explore the use of structured latent spaces to effect planning and control (e.g. \cite{srinivas2018universal,watter2015embed,banijamali2017robust,hafner2018learning}). However, existing works typically require training for a particular task on carefully curated data. In contrast, applications of variational autoencoders (VAEs) in the space of affordance-learning \cite{wu2020imagine} and quadruped locomotion \cite{mitchell2020steps} have highlighted the potential of viewing planning as run-time optimisation in pre-trained statistical models of state-space to achieve feasible spatial paths under environmental constraints.

\begin{figure}[t]
    \centering
    \includegraphics[width=\columnwidth]{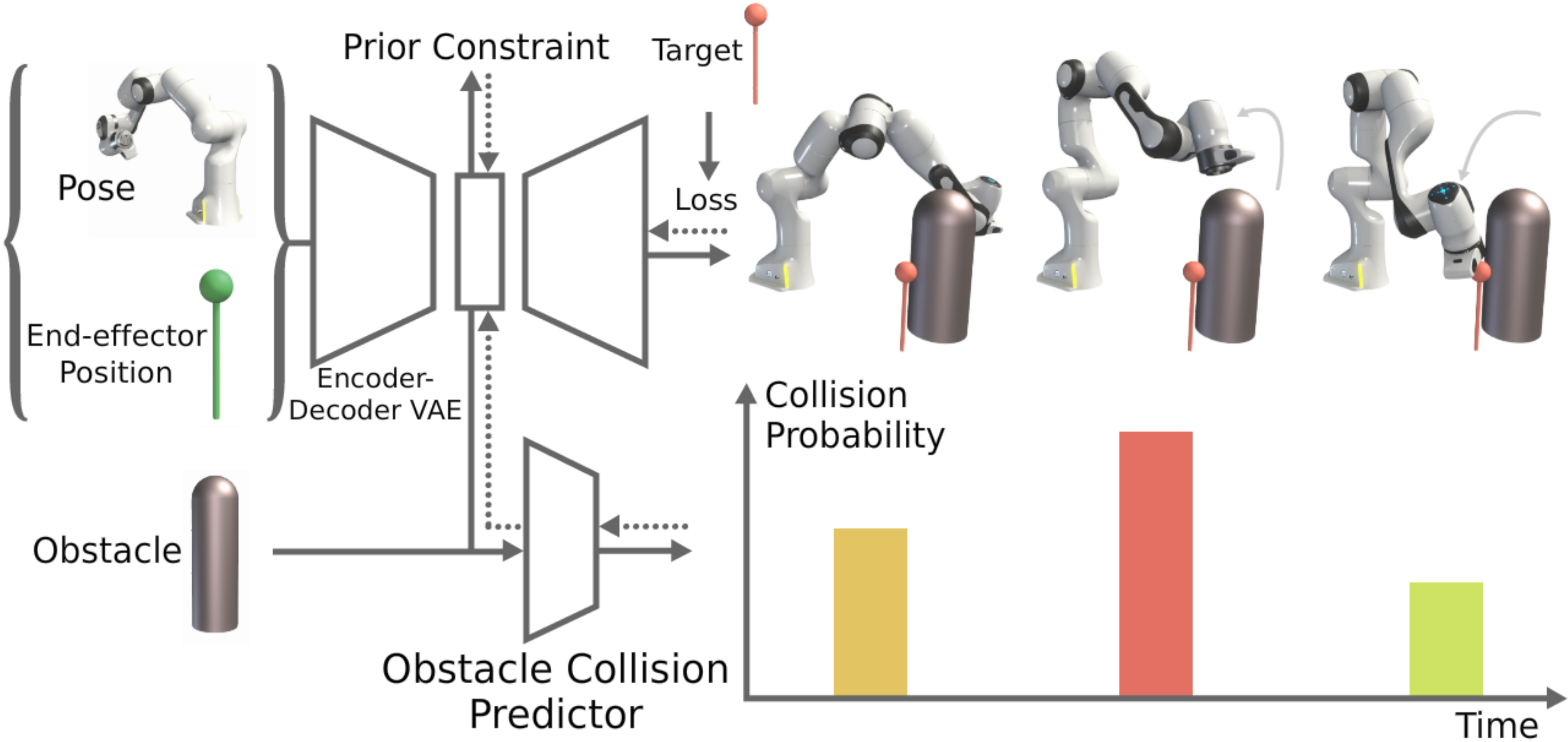}
    \caption{A VAE is trained to produce a latent representation $\bz$ of the joint states and corresponding end-effector positions and an obstacle collision predictor learns the probability of collision. Once trained, gradients through the VAE decoder and collision predictor enable optimisation in the latent space to bring the decoded end-effector position closer to the target position. Performing this optimisation iteratively with a learning rate produces a series of latent values $\{\bz_i\}_{i=1}^{T}$ that describe a joint-space path to the target that satisfies the collision constraint. 
    }
    \label{fig:teaser}
\end{figure}

Inspired by \cite{wu2020imagine} and \cite{mitchell2020steps}, in this work we explore an alternative, entirely data-driven approach to both joint-space planning and constraint satisfaction in a robot manipulation setting. In particular, our approach leverages iterative, gradient-based optimisation to produce a sequence of joint configurations by traversing the latent space of a VAE.
Training data for this model is trivially obtained as it need not be in any way task-oriented but can come from random motor-babbling on a real platform, or simply sampling valid states in simulation. In addition, \textit{performance predictors} operating on the latent space and potentially other observational data (for example, the positions of obstacles) are trained in a supervised fashion to output probabilities of certain performance-related metrics, such as whether the manipulator is in collision with an obstacle. These networks are frozen after training and are subsequently used in this gradient-based optimisation approach to planning through \textit{activation maximisation} \cite{erhan2009visualizing}, which is the process of using backpropagation through network weights to find a permutation to the network inputs that would act to bring about a desired change in the network's outputs. 

Taking this view of path planning overcomes many of the obstacles that make robotic path planning a non-trivial task:
(a) as our plans consist of states drawn from a deep generative model fit to a large dataset of feasible robot poses, and are thus approximately drawn from this data distribution, there is a very high likelihood that every state in the planned path is valid in terms of self-collisions and kinematic feasibility;
(b) by modelling joint states and end-effector positions jointly, we avoid the need to explicitly calculate inverse or forward kinematics at any stage during planning, even when the goal configuration is given in $\mathbb {R} ^{3}$ Cartesian space;
(c) by leveraging activation maximisation (AM) via gradients through performance predictors, we can enforce arbitrarily complex, potentially non-differentiable constraints that would be hard to express in direct optimisation-based planners, and might be intractably restrictive for sampling-based planners;
(d) by taking a pre-trained, data-driven approach to collision avoidance, we do not need any geometric analysis or accurate 3D models at planning time, nor indeed do we need to perform any kind of explicit collision checking, which is generally the main computational bottleneck in sampling-based planners \cite{Bialkowski2013}.

In addition to the advantages in path planning that this method offers and above and beyond related works, we introduce an additional loss on the run-time AM optimisation process which encourages the planning process to remain in areas of high likelihood according to our prior belief under the generative model. In our experiments we find that this contribution is \emph{critical} in enabling successful planning that stays in feasible state-space. 

\section{RELATED WORK}\label{sec:related_work}

Successful path planning for a robotic manipulator generally consists of producing a kinematic sequence of joint states through which the robot can actuate in order to move from a start to a goal configuration, while moving only through viable configurations. While goal positions may be specified in the same joint space as the plan, in a manipulator context it is more common for the goal position to be specified in $\mathbb {R} ^{3}$ Cartesian end-effector space, or $\mathbb {R} ^{6} $ if the \textbf{SO(3)} rotation group is included as well. Viable configurations are the intersection of feasible states for the robot - i.e. those that are within joint limits and do not result in self-collision - and collision-free states with respect to obstacles in the environment. The intersection of these defines the configuration space for the robot in a given environment. 

As analytically describing the valid configuration space is generally intractable, sampling-based methods for planning provide the ability to quickly find connected paths through valid space, by checking for the validity of individual sampled nodes. Variants of the Probabilistic Roadmap (PRM) and Rapidly-exploring Random Tree (RRT) sampling-based algorithms are widely used \cite{lavalle1998rapidly, Kavraki1996prob}, and provably asymptotically optimal variants exists in PRM*, RRT* \cite{karaman2011sampling}. These methods suffer from a trade-off between runtime and optimality: while often relatively quick to find a feasible collision-free path, they tend to employ a second, slower, stage of path optimisation to shorten the path through the application of heuristics. In the presence of restrictive constraints, both sampling- and optimisation-based planners can be very slow to find an initial feasible path \cite{Berenson2009}. 

Optimisation-based planners start from an initial path or trajectory guess and then refine it until certain costs, such as path length, are minimised, and differentiable constraints satisfied. Techniques such as CHOMP \cite{ratliff2009chomp} bridge the gap between planning and optimal control by enabling planning over path \textit{and} dynamics.
TrajOpt \cite{schulman2014motion} differs from CHOMP in the numerical optimisation method used and the method of collision checking.
The Gaussian Process Motion Planner \cite{mukadam2018continuous} leverages Gaussian process models to represent trajectories and updates them through interpolation.
Stochastic Trajectory Optimization for Motion Planning (STOMP) \cite{KalakrishnanSTOMP2011} is notable in this context as it is able to produce plans while optimising for \textit{non-differentiable} constraints , which our work enables with gradients through trained \textit{performance predictors}.

\textbf{Planning with Deep Neural Networks.}
A recent line of work has explored using deep networks to augment some or all components of conventional planning methods. Qureshi et al. \cite{Qureshi2019, Qureshi2020} train a pair of neural networks to embed point-cloud environment representations and perform single timestep planning. Iterative application of the planning network produces a path plan. Ichter and Pavone \cite{Ichter2019} learn an embedding space of observations, and use RRT \textit{in this space} with a learnt collision checker to produce path plans, but need data to learn a forward dynamics model in order to roll out the plan.

Another family of learning-based approaches to planning learn embedding spaces from high dimensional data, and learn forward dynamics models that operate on this learnt latent space. \emph{Universal Planning Networks}~\cite{srinivas2018universal} learn deterministic representations of high-dimensional data such that update steps by gradient descent correspond to the unrolling of a learned forward model. The \emph{Embed-to-Control} works \cite{watter2015embed, banijamali2018robust} employ variational inference in deep generative models in which latent-space dynamics is locally linear, a property that enables locally optimal control in these spaces. \emph{DVBF}s \cite{karl2016deep} improve on these models by relaxing the assumption that the observation space is Markovian. \emph{PlaNet}~\cite{hafner2019learning} uses a latent-space dynamics model for planning in model-based RL. However, planning in all these models tends to consist of rolling out trajectories in time, finding a promising trajectory, and executing the given actions. As such, these techniques tend to become intractable for longer time horizons, and cannot be thought of as path planning frameworks.  

A different approach is that of encoding movement primitives under the learning from demonstrations framework.
\emph{Conditional Neural Movement Primitives}~\cite{seker2019conditional} extracts prior knowledge from demonstrations and infers distributions over trajectories conditioned on the current observation.
Our approach differs in that we do not encode trajectories directly, but rather learn a probabilistic model of robot states and generate trajectories as we optimise in latent space.

\textbf{Learning Inverse Kinematics.}
In this work, by learning a joint embedding of joint angles $\ba$ and end-effector positions $\bb$, we are able to optimise for achieving an end-effector target $\bb_{\text{target}}$, while planning state sequences in \textit{joint} space.
Note that we do not care about the orientation in which the goal is reached, therefore end-effector orientation is omitted in the formulation of $\bb$.
Learning the statistical model of kinematics means we do not need to solve inverse kinematics (IK) at any point. Prior work has sought to learn solutions to IK that can cope with its ill-posed one-to-many nature for redundant manipulators \cite{bocsi2011learning, REN2020}, and to overcome the problems with analytic and numerical approaches \cite{whitney1969, goldenberg1985, wampler1986}. Ren et al.~\cite{REN2020} train a generative adversarial network to generate joint angles from end-effector positions, with the discriminator acting on the concatenation of both the input position and generated joints. This method implicitly maximises $p(\ba|\bb)$, but does not address the multimodality of the true $p(\ba|\bb)$ IK solutions. Bocsi et al.~\cite{bocsi2011learning} employed structured output learning to learn a generative model for the joint distribution of joint angles and end-effector positions. By modelling the joint instead of conditional distributions, i.e. $p(\ba, \bb)$ rather than $p(\ba|\bb)$, their model can capture the multimodal nature of IK, as one set of IK solutions $(\bb_{1}, \ba_{1})$ can be learnt without compromising the learning of another set $(\bb_{1}, \ba_{2})$.
These works are relevant in the way in which they use a learnt statistical model to capture the relationship between $\ba$ and $\bb$. 
However, we differ from prior work in that, although we follow the generative approach, we do not try to find the best $\ba$ that maximises $p(\ba, \bb)$ as is done in \cite{bocsi2011learning}, but instead plan in the latent space that decodes to valid joint configurations, producing smooth trajectories. 

While our work is partly inspired by \cite{wu2020imagine} and \cite{mitchell2020steps} in its approach, it significantly extends this prior work both in terms of method and application domain. In particular, in exploring this approach in a manipulation context we rely solely on training poses to structure the latent space. This is in contrast to \cite{mitchell2020steps}, where, in a quadruped locomotion context, structure is induced via especially designed stance labels. Like \cite{wu2020imagine}, who first proposed the use of AM for constrained optimisation in a structured latent space in the context of affordance learning, we consider environmental constraints. However, our agent operates in a significantly more complex configuration space to achieve real-world reaching and obstacle avoidance. In addition, we introduce an additional loss term that encourages the model to traverse regions of high likelihood under the learned prior over the latent variables (i.e. to stay close to the training distribution) during planning. We demonstrate that this novel loss term increases efficacy by a large margin, effectively encouraging kinematic feasibility of the plans produced.

\section{PATH PLANNING AS OPTIMISATION IN LATENT-SPACE}\label{sec:methods}

Our approach to path planning first learns a latent representation of the robot state by observing random (feasible) arm configurations. We then learn high-level performance predictors acting on this latent space as well as environment information to guide optimisation in latent space.

\subsection{Problem Formulation}

Suppose we have a dataset of joint angles and end-effector positions $\bx = \{(\ba_{i}, \bb_{i})\}_{i=1}^{m}$. We use a VAE to learn a generative latent-variable model of $\bx$.
When sampled, we expect the generative model to produce data that conform to the forward kinematics (FK) relationship. While we do not leverage the FK information at runtime, we use it during training to evaluate the \emph{sample consistency} of the generative model, i.e. how well the samples of joint angles and corresponding Cartesian end-effector positions match the actual system. 
We opt to encode $(\ba_{i}, \bb_{i})$ jointly as the information is readily available from routine robot operation and it avoids the ambiguity usually associated with mapping from the manipulator's Cartesian workspace to a valid joint configuration, thereby simplifying the inference task. To solve path planning in this approach, we use AM to iteratively backpropagate position error relative to a reaching goal into the latent space \cite{erhan2009visualizing}.
In addition, we exploit the probabilistic nature of our model by encouraging solutions to traverse regions of latent space of high likelihood under prior belief via a prior loss.
Via the decoder, each location in latent space can be decoded into a robot configuration such that trajectories in latent space, when decoded, result in sequences of robot poses.

We posit, first, that this approach will produce valid paths from an initial end-effector position to the given target position. 
Our second hypothesis is that the accuracy of reaching operation will be correlated with the sample consistency of the model. That is, if the model demonstrates a closer coupling of joint angles and Cartesian end-effector position, as defined by the analytic FK relationship, then it will produce more accurate reaching solutions via AM. We will demonstrate how the approach can be extended to deal with reaching tasks while avoiding obstacles. One strength of this approach is the conceptual ease with which additional constraints can be added.

\subsection{Learning a Latent Representation of Robot State}

Our aim is to learn a generative model of $\bx$.
This can be accomplished with a variational autoencoder (VAE) \cite{kingma2013auto, rezende2014stochastic}, which defines an encoder $q_\phi(\mathbf{z}\mid\mathbf{x})$ and decoder $p_\theta(\mathbf{x}\mid\mathbf{z})$, where $\bz$ is a learned latent representation.
To train the VAE, we would like to maximise the evidence, $p_\theta(\bx) = \int p_\theta(\bx \mid \bz) p_\theta(\bz) d\bz$, which is generally intractable. A common alternative therefore is to maximise the evidence lower bound (ELBO), where $\mathcal{L}^{\text{ELBO}} \leq \log p(\bx)$: 
\begin{align}\label{eq:elbo}
    \mathcal{L}^\text{ELBO} = 
    \underbrace{\mathbb{E}_{\mathbf{z}\sim q_\phi(\mathbf{z} \mid \mathbf{x})} \log p_\theta (\mathbf{x} \mid \mathbf{z})}_{\text{Reconstruction Accuracy}}
     - 
    \underbrace{D_{\text{KL}}\left[q_\phi(\mathbf{z} \mid \mathbf{x}) \mid\mid p(\mathbf{z})\right]}_{\text{KL Term}} 
\end{align}
To trade off between reconstruction accuracy and the KL term, a $\beta$ hyperparameter is often added to the ELBO formulation \cite{higgins2017beta}. Rather than setting this hyperparameter manually \cite{higgins2017beta}, we adopt an alternative, dynamic GECO approach \cite{rezende2018taming}.
The GECO objective formulates the ELBO loss as a constrained optimisation problem, using a Lagrange multiplier $\lambda$, such that
\begin{align}\label{eq:geco}
    \mathcal{L}^\text{GECO} =
    \underbrace{-D_{\text{KL}}[q_\phi (\mathbf{z} \mid \mathbf{x}) \mid\mid p(\mathbf{z})]}_{\text{KL Term}}
    + \lambda 
    \underbrace{\mathbb{E}_{\mathbf{z}\sim q_\phi (\mathbf{z} \mid \mathbf{x})} \left[ \mathcal{C} \left( \mathbf{x}, \hat{\mathbf{x}} \right) \right]}_{\text{Reconstruction Error Constraint}},
\end{align}
where $\hat{\mathbf{x}}$ is the reconstruction of $\mathbf{x}$ through the VAE.
The reader is referred to \cite{rezende2018taming} for implementation details on how $\lambda$ is updated.
The Lagrangian optimises the KL divergence subject to $\mathbb{E}_\textbf{z}\left[ \mathcal{C}\left(\mathbf{x},\hat{\mathbf{x}} \right) \right] \leq 0$, for a given constraint function $\mathcal{C}$. The constraint typically models an upper bound on a predefined reconstruction error (e.g.\ an $L_2$ loss):
\begin{align}
    \mathcal{C} \left( \mathbf{x}, \hat{\mathbf{x}} \right) = 
    \norm{\mathbf{x} - \hat{\mathbf{x}}}_2 - \tau
\label{eq:geco_recon_target}
\end{align}
Although the GECO formulation still contains a hyperparameter, $\tau \geq 0$, this represents an interpretable quantity: an upper bound on the reconstruction error. In practice, this is easier to work with than tuning the $\beta$ hyperparameter in the latent space, which is difficult to interpret.
VAEs in our experiments are trained by optimising the GECO objective with the $L_2$ reconstruction loss.

\subsection{Activation Maximisation for Path Planning Under a Prior Loss} \label{sec:am_for_plan}

Given a target position $\bbb$, the aim is to produce a sequence of joint configurations $(\ba_0, \dots, \ba_T)$ that drive the robot's end-effector from its initial position $\bb_0$ to an end position $\bb_T$ within a distance tolerance $\norm{\bb_T, \bbb}_2 < \gamma$. This can be achieved in the probabilistic model through the iterative use of AM~\cite{erhan2009visualizing}.

Let the initial $\bx_0$ be encoded such that the corresponding latent configuration $\bz_0$ is drawn from the posterior. Decoding $\bz_0$ then gives rise to $\hat{\bx}_0 = \{\hat{\ba}_0, \hat{\bb}_0\}$.
More generally, $\hat{\bx} = \{\hat{\ba}, \hat{\bb}\}$.
Let $\norm{\hat{\bb}_0, \bbb}_2$ denote the Euclidean distance between $\hat{\bb}_0$ and $\bbb$, then we can compute an $L_2$ loss that we backpropagate through the decoder $p_\theta(\bb, \ba \mid \bz)$. However, rather than update the network weights, we use AM to update the latent vector. In particular, given the AM objective, latent representations are updated iteratively in the following way, where $\alpha_{\text{AM}}$ is the learning rate and $\nabla \mathcal{L}^{\text{AM}}$ is the gradient of the AM loss with respect to the input $\bz$:
\begin{align}
    {\bz}_{t+1} = {\bz}_t - \alpha_{\text{AM}} \nabla \mathcal{L}^{\text{AM}},  \qquad
    \mathcal{L}^{\text{AM}} = 
    \underbrace{\norm{\hat{\bb}, \bb_{\text{target}}}_2}_{\text{Target Loss}}
\end{align}
This produces a progression of latent representations $(\bz_1, \dots, \bz_T)$, which continues for a set number of $T$ steps. Through the decoder, these latent representations can be mapped to joint configurations $(\ba_1, \dots, \ba_T)$. If the kinematics relationships represented by the decoder network are valid, and a sufficient number of steps $T$ are taken,
then we expect the final joint angle configuration $\ba_T$ to correspond to a new end-effector position $\bb_T$ such that $\norm{\bb_T, \bb_{\text{target}}}_2 < \gamma$.
Starting with the initial position, the sequence of decoded end-effector positions represents a spatial path $(\bb_0, \dots, \bb_T)$.

Without modification, AM may often drive the values $\bz$ into parts of the latent space that have not been seen during training.
Decoding these latent representations can lead to poor $(\ba, \bb)$ pairs that are inconsistent with the desired kinematics. To encourage the optimisation to traverse regions in which the model is well defined (i.e. to stay as close to the training distribution as possible) we introduce an additional loss term to the AM objective consisting of the likelihood of the current latent representation under its prior $p(\bz)$, such that 
\begin{align}
    \mathcal{L}^{\text{AM}} = 
    \underbrace{\norm{\hat{\bb}, \bb_{\text{target}}}_2}_{\text{Target Loss}} +
    \lambda_{\text{prior}} \underbrace{(-\log p(\bz))}_{\text{Prior Loss}}
\label{eq:am-loss}
\end{align}
This encourages the reconstructed joint configurations to remain valid. Again, $\lambda_{\text{prior}}$ is tuned automatically during training using a GECO formulation, by selecting an upper bound on the prior loss.

\subsection{Obstacle Avoidance via Performance Predictors} \label{sec:am_obs_avoid}

A key requirement for path planning is obstacle avoidance. In our framework this is effected by a binary classifier predicting whether the current arm configuration, as represented in latent space, is in collision with an obstacle. By back-propagating gradients forcing the collision response of this classifier to zero we effectively drive the robot away from obstacles. 
The classifier is trained using a binary cross-entropy (BCE) loss while the VAE weights remain frozen.

When performing AM in the case of obstacle avoidance, we add an obstacle loss term from BCE to the AM loss in Eq.~\ref{eq:am-loss}.

\begin{equation}
\begin{aligned}
    \mathcal{L}^{AM} = &
    \underbrace{\norm{\hat{\bb}, \bb_{\text{target}}}_2}_{\text{Target Loss}} +
    \lambda_{\text{prior}} \underbrace{(-\log p(\bz))}_{\text{Prior loss}} \\
    & +
    \lambda_{\text{obs}} \underbrace{\sum_i (-\log (1 - p_\vartheta(\bz, \bo_i)))}_{\text{Obstacle loss}},
\label{eq:am-loss-obstacle-avoidance}
\end{aligned}
\end{equation}
where $\lambda_{\text{prior}}$ and $\lambda_{\text{obs}}$ are tuned jointly using GECO with mutliple constraints. Avoidance of multiple obstacles can be achieved by repeatedly deploying the same classifier and adding the resulting gradients into the optimisation. The ease with which multiple constraints can be expressed and enforced is an explicit strength of this approach.

\subsection{Model Selection Through Sample Consistency} \label{sec:sample_cons}

While the downstream performance we seek from our models is better path planning, this is not continuously measurable during training. For VAE model selection and hyperparameter tuning, we consider three metrics as predictors of path planning success: (a) the data reconstruction loss $\|\hat{\bb}-\bb\|_{2}+\|\hat{\ba}-\ba\|_{2}$, (b) ELBO (Eq.~\ref{eq:elbo}) and (c) kinematic sample consistency, which we define as
\begin{equation}
    \delta=\|\hat{e}-\mathrm{FK}(\hat{\boldsymbol{q}})\|_{2}
\end{equation}
This \emph{sample consistency error} $\delta$ is the Euclidean distance between the reconstructed end-effector position $\hat{\bb}$ and the true forward kinematics (FK) solution for the reconstructed joint angles $\hat{\ba}$. We find that high sample consistency is a better predictor of a model's downstream planning performance than the more traditional ELBO loss alone (Fig.~\ref{fig:success_vs_distance_threshold_vs_sample_consistency} right). 

\section{IMPLEMENTATION DETAILS}
\label{sec:panda-experiments}
This section provides details on model architecture, model training and planning, using a 7-Dof Emika Franka Panda arm.

\subsection{Architecture Details}
\label{subsec:3d-architecture}
The VAE architecture comprises of an encoder and a decoder. The encoder takes as input $\bx=\{\ba,\bb\}$ and outputs the mean $\bmu$ and variance $\bsig$ of the posterior distribution $q_\phi(\mathbf{z} \mid \mathbf{x})$. The latent encoding $\bz$ is then obtained using the reparameterisation trick \cite{kingma2013auto}. A multivariate isotropic Gaussian prior is imposed on the latent space. The decoder takes as input the latent sample $\bz$ and outputs the reconstruction $\hat{\bx}=\{\hat{\ba},\hat{\bb}\}$.
The obstacle collision classifier takes $\{\bz, \bo=\{x,y,h,r\}\}$ (xy coordinates, height, radius of the cylinder) as input and has a single output logit, which when passed through a sigmoid function gives the predicted probability of collision.
The encoder, decoder and obstacle collision classifier each contains four fully connected hidden layers of 2048 units, but differ in input and output layers, as shown in Table \ref{table:nnarchitecture}.

\def\arraystretch{1.25}

\begin{table}[h!]
\setlength{\tabcolsep}{2.5pt}

\begin{center}
\begin{tabular}{l|ccc}
\toprule
& VAE Encoder & VAE Decoder & Collision Classifier \\
\hline
Input Dimension & 10 & 7 & 11 \\ \hline
Output Dimension & 7$\times$2 & 10 & 1 \\ \hline
No. of Hidden Layers & 4 & 4 & 4 \\ \hline
Units per Hidden Layer & 2048 & 2048 & 2048 \\ \hline
Hidden Layer Activation & ELU & ELU & ELU \\
\bottomrule
\end{tabular}
\end{center}
\vskip 0.1in
\caption{The architecture for the encoder, the decoder, and the obstacle collision classifier. The VAE encoder takes input $\{\ba,\bb\}$, where $dim(\ba)=7$, $dim(\bb)=3$, and outputs ${\bmu,\bsig}$, where $dim(\bmu)=dim(\bsig)=7$. The VAE decoder takes input $\bz$ where $dim(\bz)=7$, and outputs reconstruction which is of the same dimension as the input. The collision classifier takes input $\{\bz, \bo\}$, where $dim(\bz)=7$, $dim(\bo)=4$.
}
\label{table:nnarchitecture}
\end{table}

\subsection{Training Data Generation}
In this evaluation, we consider cylindrical objects\footnote{Our approach readily extends to other obstacle geometries, extending to observation space.}, which are easily represented in state space as tuples $\{\ba, \bb, \bo, \textbf{c}\}$, where $\ba=(\theta_1,...,\theta_7)$ represents the robot joint configurations; $\bb=(e_1, e_2, e_3)$ the end-effector coordinates; $\bo=(x,y,h,r)$ the obstacle coordinates, height and radius; and $\textbf{c} \in \{0, 1\}$ the binary collision label.
Joint configurations $\ba$ are sampled uniformly within the joint limits. We take the modified Denavit–Hartenberg parameters in the Panda arm documentation to characterise the forward kinematics relationship, $\bb=\FK(\ba)$.
To generate the position of the obstacles, for each obstacle, we sample a distance to origin $L$, an angle $\theta_{obs}$ in $[0, 2\pi)$ uniformly and set $x=L\, \cos(\theta_{obs})$, $y=L\, \sin(\theta_{obs})$.
MoveIt's planning scene interface is used to check whether the arm is in self-collision or in collision with the table; joint configurations that are in such collisions are discarded. We also use MoveIt's planning scene interface to label collision with obstacles in the training data. The dataset contains an equal number of samples in and not in collision with the obstacles. In total, the dataset contains 100k data points, of which 80k are used for training and 20k for validation.

\subsection{Obstacle Scenario Generation}
In our experiments, scenarios are generated by sampling a given number of obstacles and two sets of joint angles -- one for the initial robot configuration and another for the target position.
The joint angle samples for the target are only used to compute the target position through the FK model of the Panda arm that we characterised and are not known to the planner. 
The obstacles and the target are generated while ensuring that there is at least a feasible set of joint angles reaching the target without collision with the obstacles.
The first obstacle is sampled between the initial end-effector position and the target position. Subsequent obstacles are either sampled randomly or sampled between the initial end-effector position and the target position, with a probability of 50\% each. The model is evaluated on scenarios in which it would collide with the obstacles if the obstacle loss term was not added to the total loss in the AM objective function.

\subsection{Training Details}
The input values to the VAE and the collision classifier are standardised, and the output values de-standardised, according to the mean and the standard deviation of the training data.
The model is trained using a batch size of 256 for 16,000 epochs using the Adam optimiser~\cite{kingma2014adam}.
To select hyperparameters, a grid search is run on the following values: number of hidden layers, units per layer, latent dimension, GECO reconstruction target $\tau$ (Eq.~\ref{eq:geco_recon_target}), VAE learning rate, and GECO learning rate.

\subsection{Planning Details}

Planning is achieved by applying activation maximisation in the latent space, as outlined in Algorithm~\ref{algo:am_obs_avoidance}. 
\begin{algorithm}[h]
\SetAlgoLined
initialise path buffer $X$\;
initialise $\lambda_{prior}$ and $\lambda_{obs}$\;
infer latent representation of initial configuration $\mathbf{z}_0\sim q_\phi(\mathbf{z} \mid \mathbf{x}=\mathbf{x}_0)$\;
 \For {\textup{each time step t = 0,1,...,T}} {
  decode latent encoding to state space $\hat{\mathbf{x}}_t \sim p_\theta (\mathbf{x} \mid \mathbf{z}=\mathbf{z}_t)$\;
  save states to path $X_t = \hat{\mathbf{x}}_t$\;
  \If{$d(\bb_t, \bbb) < \gamma$}{
    break\;
  }
  compute loss terms $\norm{\hat{\bb}, \bb_{\text{target}}}_2$, $-\log p(\bz)$, and $\sum_i (-\log (1 - p_\vartheta(\bz, \bo_i)))$\;
  $\lambda^{t+1}_{\text{prior}}$ = Update ($\lambda^t_{\text{prior}}$)\;
  $\lambda^{t+1}_{\text{obs}}$ = Update ($\lambda^t_{\text{obs}}$) \;
  compute $\mathcal{L}^{AM}_t = 
    \norm{\hat{\bb}, \bb_{\text{target}}}_2 +
    \lambda^{t+1}_{\text{prior}} (-\log p(\bz)) +
    \lambda^{t+1}_{\text{obs}} \sum_i (-\log (1 - p_\vartheta(\bz, \bo_i)))$\;
  update ${\bz}_{t+1} = {\bz}_t - \alpha_{\text{AM}} \nabla \mathcal{L}^{\text{AM}}_t$\;
 }
 \caption{Activation Maximisation for Obstacle Avoidance}
\label{algo:am_obs_avoidance}
\end{algorithm}

\section{RESULTS}
\label{sec:panda-results}

We evaluate our approach in the context of a set of robot reaching tasks, described below, using a simulated Panda arm. We further demonstrate that the approach can be deployed on a physical Panda arm.

\subsection{Path Planning for Target Reaching}
\begin{figure}[bht]
\includegraphics[width=.495\columnwidth]{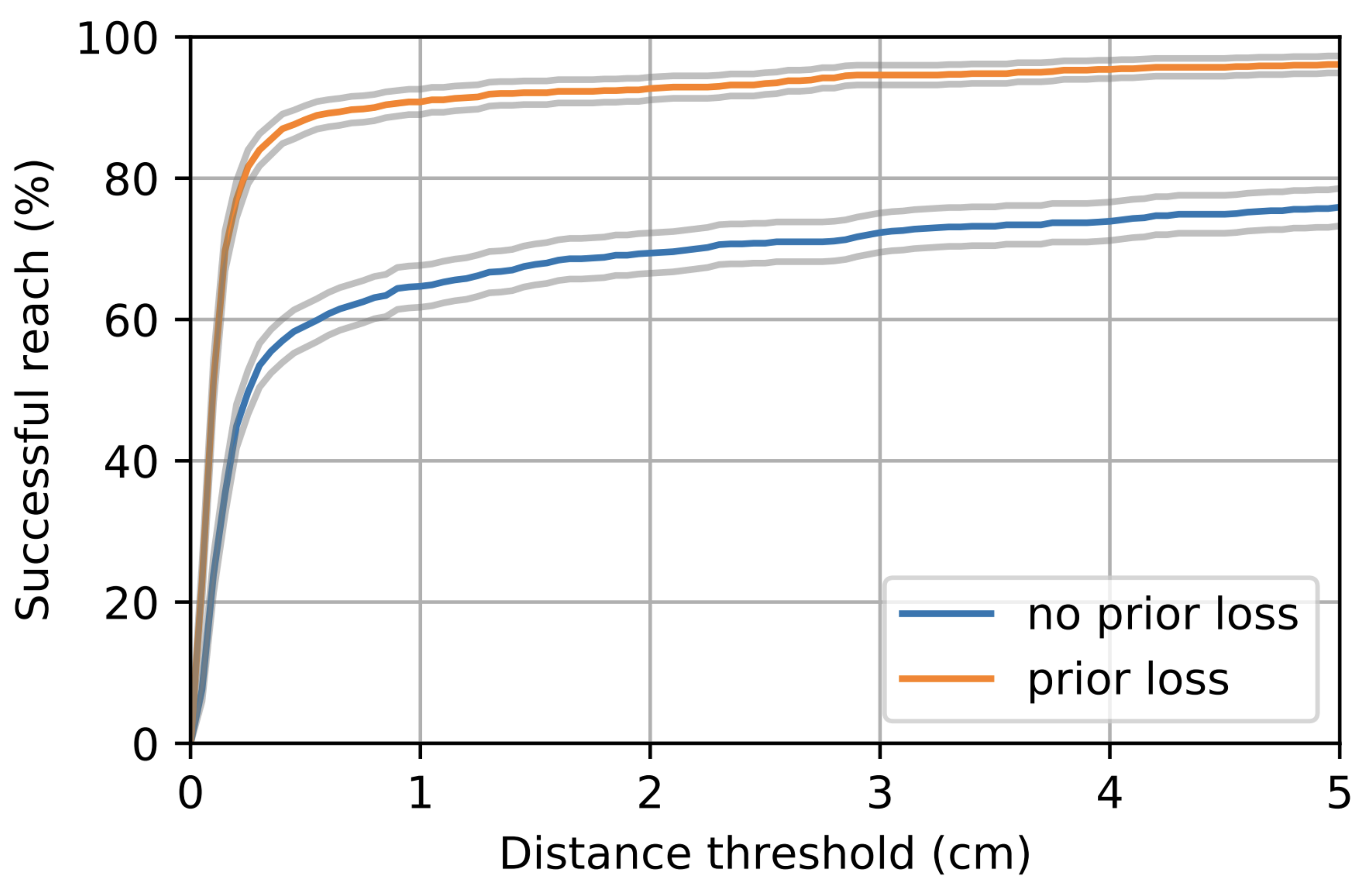}
\includegraphics[width=.495\columnwidth]{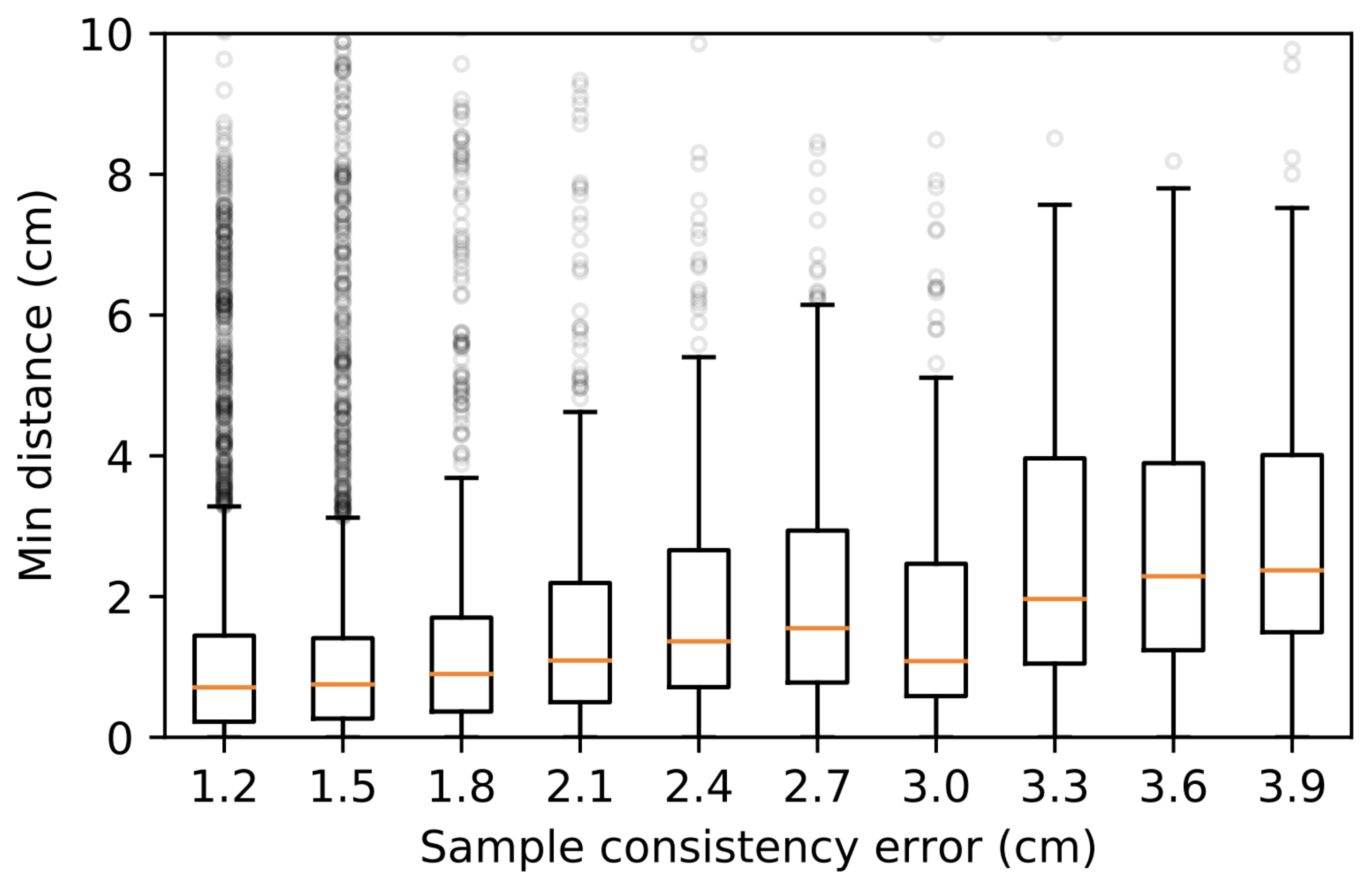}
\caption{Left: Target reaching success vs reaching distance threshold, evaluated on 1,000 scenarios. Grey lines are the 95\% confidence interval of Wilson score~\cite{wilson1927probable}. Adding prior loss in AM objective function improves reaching success rate. Right: Minimum distance between the end-effector and the target vs sample consistency error. Lower sample consistency error leads to better target reaching.}
\label{fig:success_vs_distance_threshold_vs_sample_consistency}
\end{figure}

\begin{figure}[bt]
    \centering
    \includegraphics[width=\columnwidth]{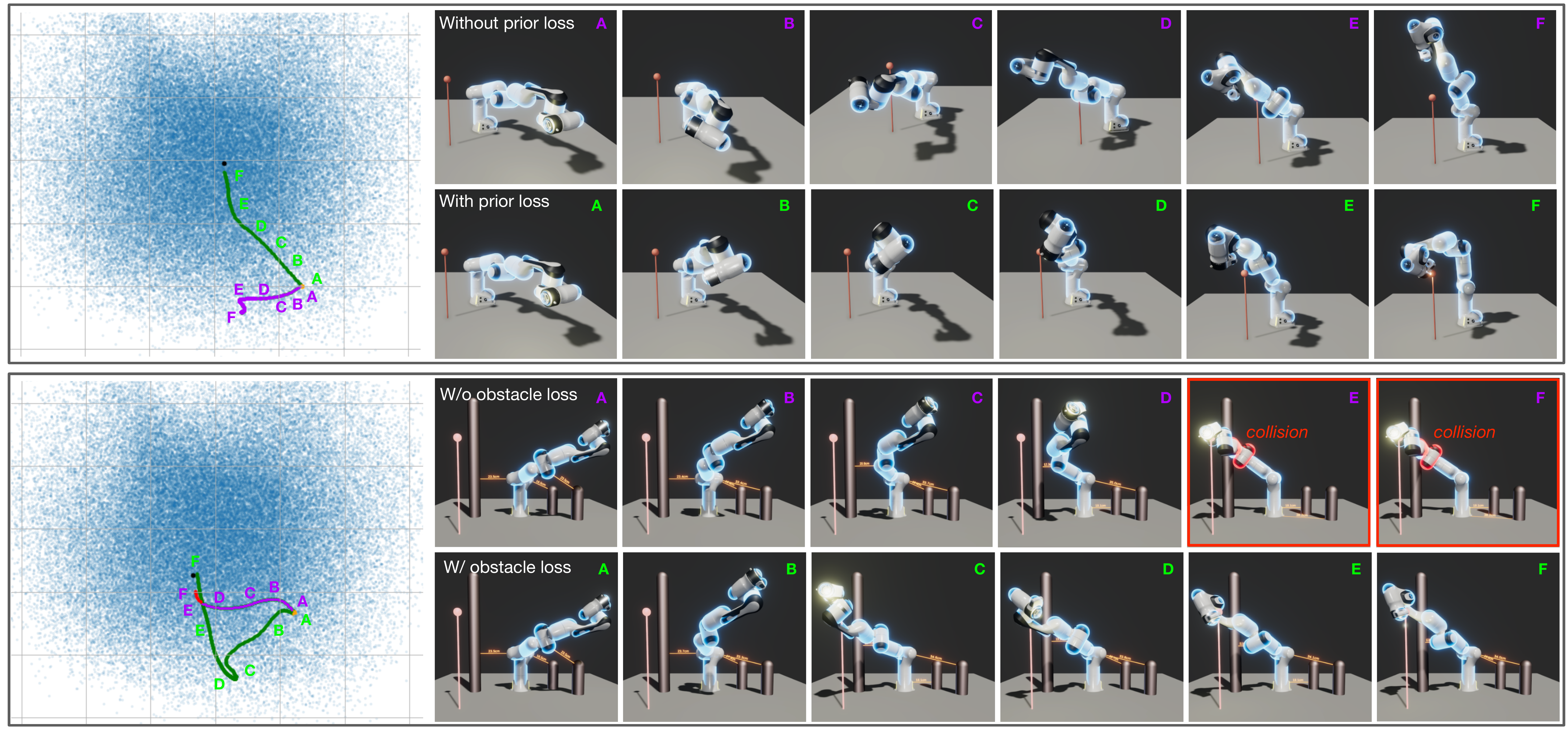}
    \caption{We project the latent space down to 2D via PCA to visualise AM with and without prior loss (top) / with and without obstacle loss (bottom). The blue region is the encoding of the training distribution. The green and the purple curves are the robot trajectories from AM. The black dot is the latent representation of the target joint angles and coordinates. In the case of no prior loss, the encoding of the robot initial configuration lies in the trusted region, but drifts to its boundary as we perform gradient descent, which decodes to a meaningless output. In the case of no obstacle loss, the robot collides with an obstacle. The link in collision is shown in red.}
    \label{fig:pca-prior-obstacle}
\end{figure}

Before extending to path planning with additional constraints, we explore the ability of iterative AM as described in Section~\ref{sec:am_for_plan} to produce a path plan for goal reaching in free space.
We sample 1,000 start and goal configurations for the robot, with an initial joint position $\ba_{1}$ and a goal $\bbb$ in $\mathbb {R} ^{3}$. Results of our method are shown in Fig.~\ref{fig:success_vs_distance_threshold_vs_sample_consistency} (left), where we quantify planning success rates at different distance thresholds. We find that the addition of the \textit{prior loss} (Eq.~\ref{eq:am-loss}) to the AM objective is instrumental in improving success rates, while when we optimise AM for reducing distance to goal with no additional constraint, we observe more frequent
infeasible state reconstructions $\hat{\ba}$ in the path plans (Fig.~\ref{fig:pca-prior-obstacle} top). With the prior loss, over 90\% of the planning scenes are solved to within a 5mm threshold of the goal. 

\subsection{Obstacle Avoidance}

\begin{figure}[tb]
    \centering
    \includegraphics[width=\columnwidth]{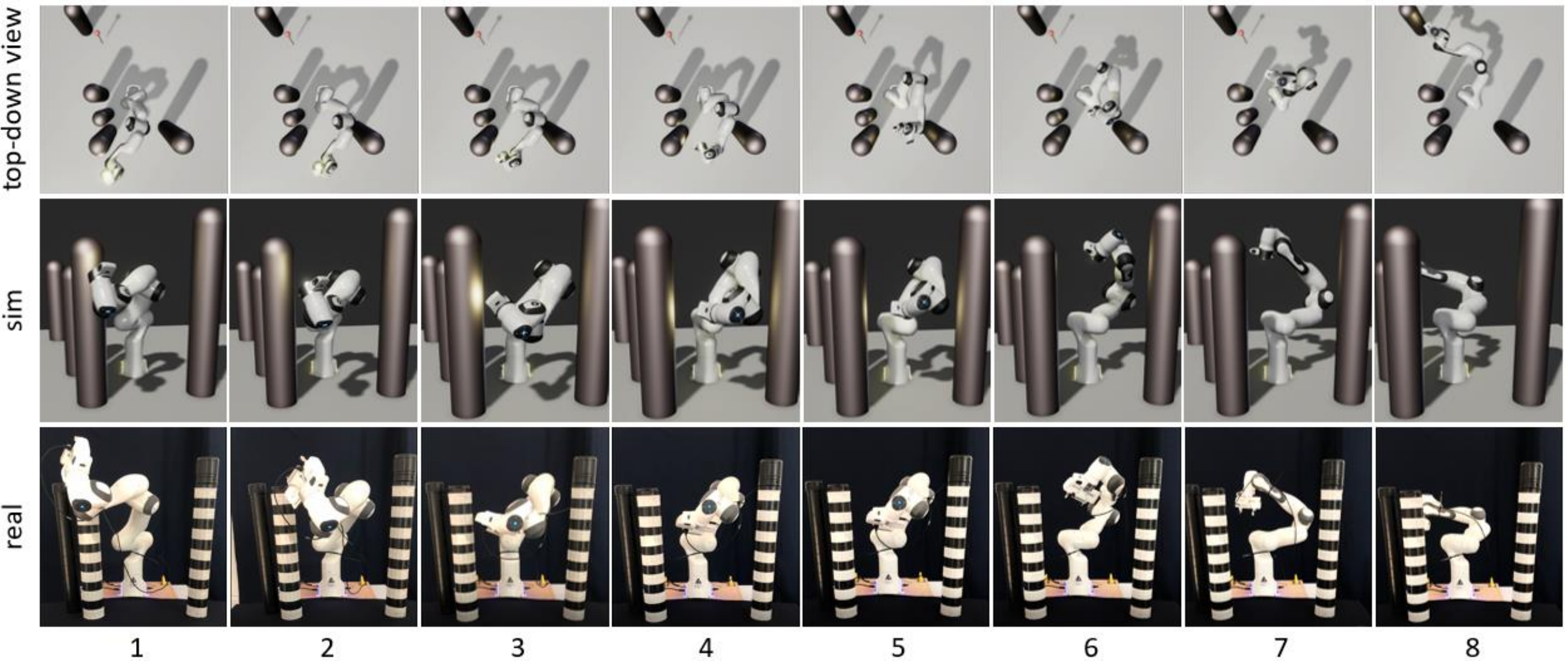}
    \caption{Real-world experiments: a rollout of a trajectory using LSPP. Our latent-space approach operates in state space and therefore trivially transfers to the real world.}
    \label{fig:panda_obs_rollout}
\end{figure}

To demonstrate the effect of adding obstacle loss to accomplish obstacle avoidance, we show qualitative results both in simulation and in the real world in Fig.~\ref{fig:pca-prior-obstacle} (bottom) and Fig.~\ref{fig:panda_obs_rollout}.

To evaluate the efficacy of our approach in finding feasible plans in the presence of obstacles we generate 1,000 scenarios for each of one to five cylindrical obstacles. We compare our approach of latent-space path planning (LSPP) to eight planners in widespread use: Potential Field~\cite{khatib1986real, flacco2012depth}, RRTConnect~\cite{kuffner2000rrt}, LBKPIECE~\cite{csucan2009kinodynamic}, RRT*~\cite{karaman2011sampling}, LazyPRM*~\cite{bohlin2000path}, FMT*~\cite{janson2015fast}, BIT*~\cite{gammell2020batch} and CHOMP~\cite{ratliff2009chomp}.
The artificial potential field baseline is a classical local collision avoidance method using Jacobian pseudo-inverse to reduce the error in end-effector position while avoiding obstacles through the use of virtual repulsive forces.
It is adapted from \cite{flacco2012depth}, but instead of using the depth-space concept to estimate the distances between the robot and the obstacles, it has direct access to the obstacle configurations to generate the repulsive vectors. A grid search is conducted on the hyperparameters (Eq. 7 in \cite{flacco2012depth}) to optimise for the overall success rate.
For all other baselines we use the default parameters from their MoveIt OMPL and CHOMP library implementations~\cite{chitta2012moveit, sucan2012open}. For RRT, LazyPRM* and BIT*, we keep the default planning time of 5 seconds.
CHOMP uses a linear initialisation from start to goal position and optimises locally, such that it provides a fair comparison for local planners.
Quantitative results are shown in Table~\ref{table:obstacle-avoidance-success-rate}.

For LSPP, a grid search is conducted on the GECO target (Eq.~\ref{eq:am-loss-obstacle-avoidance}), GECO smoothing factor and GECO learning rate for the obstacle loss term to optimise for the overall success rate.
Across all methods, a run is considered a success if the robot reaches the target within a distance threshold of 1cm and without colliding with obstacles.

\begin{table*}[bt]

\setlength{\tabcolsep}{3pt}

\centering
\textbf{Planning success rate {[}\%{]}}
\vspace{3pt}

\begin{tabular}{p{55pt}||C{58pt}C{58pt}C{58pt}C{58pt}C{58pt}}
\#obstacles & 1 & 2 & 3 & 4 & 5 \\ \hline \hline
LSPP (ours) & \textbf{85.8 $\pm$ 2.2} & \textbf{59.4 $\pm$ 3.0} & 38.2 $\pm$ 3.0 & 25.0 $\pm$ 2.7 & 15.7 $\pm$ 2.3 \\ \hline
Potential Field & 34.2 $\pm$ 2.9 & 20.5 $\pm$ 2.3 & 15.7 $\pm$ 2.3 & 11.4 $\pm$ 2.0 & \, 5.3 $\pm$ 1.4 \\
RRTConnect & 84.9 $\pm$ 2.2 & 58.8 $\pm$ 3.1 & 47.7 $\pm$ 3.1 & 34.5 $\pm$ 2.9 & 26.8 $\pm$ 2.7 \\
LBKPIECE & 82.9 $\pm$ 2.3 & 57.8 $\pm$ 3.1 & \textbf{49.1 $\pm$ 3.1} & 32.5 $\pm$ 2.9 & 25.3 $\pm$ 2.7 \\
RRT* & 85.0 $\pm$ 2.2 & 58.1 $\pm$ 3.1 & 47.7 $\pm$ 3.1 & 33.2 $\pm$ 2.9 & 25.9 $\pm$ 2.7 \\
LazyPRM* & 82.3 $\pm$ 2.4 & 57.5 $\pm$ 3.1 & 47.2 $\pm$ 3.1  & 33.2 $\pm$ 2.9 & 25.4 $\pm$ 2.7 \\
FMT* & 66.5 $\pm$ 2.9 & 52.7 $\pm$ 3.1 & 37.2 $\pm$ 3.0  & 29.8 $\pm$ 2.8 & 15.7 $\pm$ 2.3 \\
BIT* & 85.7 $\pm$ 2.2 & 58.0 $\pm$ 3.1 & 48.3 $\pm$ 3.1  & \textbf{34.8 $\pm$ 3.0} & \textbf{26.9 $\pm$ 2.7} \\
CHOMP & 80.0 $\pm$ 2.5 & 56.1 $\pm$ 3.1 & 37.0 $\pm$ 3.0 & 25.1 $\pm$ 2.7 & 16.2 $\pm$ 2.3
\end{tabular}

\vspace{3pt}

\centering
\textbf{Planning time [ms]}
\vspace{3pt}

\begin{tabular}{p{55pt}||C{58pt}C{58pt}C{58pt}C{58pt}C{58pt}}
\#obstacles & 1 & 2 & 3 & 4 & 5 \\ \hline \hline
LSPP (ours) & \, 179.8 $\pm$ \, 85.1 & \, 185.5 $\pm$ \, 90.8 & \, 189.8 $\pm$ \, 91.5 & \textbf{\, 191.9 $\pm$ \, 92.0} & \textbf{\, 201.0 $\pm$ \, 98.2} \\ \hline
Potential Field & 1973.5 $\pm$ 296.4 & 2048.2 $\pm$ 313.6 & 2104.1 $\pm$ 320.6 & 2125.8 $\pm$ 327.2  & 2148.4 $\pm$ 319.7 \\
RRTConnect & \textbf{\, 128.3 $\pm$ 254.0} & \textbf{\, 150.5 $\pm$ 330.0} & \textbf{\, 180.9 $\pm$ 390.3}  & \, 195.9 $\pm$ 344.2  & \, 231.9 $\pm$ 252.8 \\
LBKPIECE & \, 401.7 $\pm$ 400.1 & \, 437.0 $\pm$ 455.9 & \, 526.8 $\pm$ 601.4 & \, 539.1 $\pm$ 451.2 & \, 561.6 $\pm$ 330.5 \\
FMT* & \, 877.8 $\pm$ 215.3 & \, 887.8 $\pm$ 199.4 & \, 872.6 $\pm$ 241.0 & \, 807.4 $\pm$ 206.0 & \, 820.9 $\pm$ 225.8 \\
CHOMP & \, 526.2 $\pm$ 308.7 & \, 628.3 $\pm$ 286.5 & \, 745.2 $\pm$ 316.1 & \, 792.4 $\pm$ 280.7 & \, 873.4 $\pm$ 412.5
\end{tabular}

\vspace{3pt}

\centering
\textbf{Path length}
\vspace{3pt}

\begin{tabular}{p{55pt}||C{58pt}C{58pt}C{58pt}C{58pt}C{58pt}}
\#obstacles & 1 & 2 & 3 & 4 & 5 \\ \hline \hline
LSPP (ours) & \textbf{1.52 $\pm$ 0.36} & 1.51 $\pm$ 0.34 & \textbf{1.47 $\pm$ 0.30} & \textbf{1.50 $\pm$ 0.31} & 1.48 $\pm$ 0.27 \\ \hline
Potential Field & 1.54 $\pm$ 0.44 & 1.57 $\pm$ 0.35 & 1.54 $\pm$ 0.37 & 1.53 $\pm$ 0.36 & 1.53 $\pm$ 0.37 \\
RRTConnect & 2.33 $\pm$ 1.23 & 2.25 $\pm$ 1.05 & 2.24 $\pm$ 1.13 & 2.12 $\pm$ 1.05 & 2.15 $\pm$ 1.14 \\
LBKPIECE & 2.27 $\pm$ 1.14  & 2.26 $\pm$ 1.25 & 2.16 $\pm$ 0.99 & 2.07 $\pm$ 1.04 & 1.94 $\pm$ 0.99 \\
RRT* & 1.53 $\pm$ 0.93 & \textbf{1.50 $\pm$ 0.67} & 1.48 $\pm$ 0.67 &  \textbf{1.50 $\pm$ 0.83} &  \textbf{1.47 $\pm$ 0.57} \\
LazyPRM* & 2.20 $\pm$ 1.11 & 2.18 $\pm$ 1.22 & 2.13 $\pm$ 1.07 & 2.03 $\pm$ 0.97 & 1.96 $\pm$ 0.87 \\
FMT* & 2.30 $\pm$ 1.07 & 2.01 $\pm$ 0.84 & 2.04 $\pm$ 0.66 & 1.94 $\pm$ 0.69 & 1.91 $\pm$ 0.50 \\
BIT* & 2.13 $\pm$ 1.18 & 1.94 $\pm$ 0.73 & 1.87 $\pm$ 0.56 & 2.06 $\pm$ 0.74 & 1.98 $\pm$ 0.77 \\
CHOMP & 2.28 $\pm$ 1.23  & 2.27 $\pm$ 1.20 & 2.25 $\pm$ 1.15 & 2.16 $\pm$ 1.12 & 1.98 $\pm$ 0.93
\end{tabular}

\bigskip

\caption{Comparison of performance of our latent-space path planning (LSPP) and baseline motion planning algorithms. For each number of obstacles, the experiments are run on a test dataset of 1,000 scenarios. The values are displayed with a 95\% confidence interval (Wilson score~\cite{wilson1927probable} for planning success rate and standard deviation for planning time and path length).
The planning time of RRT*, LazyPRM* and BIT* are omitted since they operate with a fixed time budget of 5 seconds.}
\label{table:obstacle-avoidance-success-rate}
\end{table*}

In terms of planning success rate, LSPP performs commensurate to the baselines in the case of one and two obstacles, but suffers a performance drop when more obstacles are present.
This performance drop is expected and can also be observed in CHOMP, another optimisation-based planner.
Our scenario generation process does not ensure there exists a feasible solution to a particular scenario. The success rates are therefore only indicative of relative performance. However, RRTConnect, RRT*, FMT*and BIT* serve as useful calibration as they are probabilistically complete, ensuring a solution will be found if one exists, given sufficient runtime.
There are a number of factors which influence LSPP performance. There exists an inherent tension due to the AM objective between reaching a goal and avoiding obstacles. This is, in effect, regulated by the GECO parameters.
As LSPP is inherently a gradient-based optimisation method it is subject to local minima.
Empirically, this happens more often as the number of obstacles increases, but could potentially be handled by adding a stochastic recovery strategy or a post processor.
In addition, the optimisation can be misguided either by a failure in the obstacle classifier or due to low sample consistency. 

Overall LSPP's average planning time is commensurate with that of RRTConnect whereas it significantly outperforms Potential Field, LBKPIECE, CHOMP and FMT*. We note also that LSPP exhibits consistently lower variances in planning time than the baselines. In LSPP, each additional obstacle requires an extra forward and backward pass of the collision predictor, and thus planning time increases linearly with obstacles. However, in these experiments this remains a negligible effect on the overall LSPP time.

The path length is normalised by dividing the actual length of the planned path by the Euclidean distance between the initial end-effector position and the target position to ensure a fairer comparison among different scenarios.
It should be noted that the cost functions in OMPL minimise joint space path length, and a shortest path in joint space does not necessarily translate into a shortest path in Cartesian space.
RRT* is an asymptotically optimal algorithm, thus it is not surprising that it finds near optimal paths. Nevertheless, LSPP outperforms most of the other baselines.

The artificial potential field baseline is widely used due to its simplicity and serves as a useful comparison for local collision avoidance methods. It is in spirit most similar to LSPP, subject to local minima, and neither of them has theoretical guarantees. However, it is not directly comparable as it assumes access to the FK relationship to compute the Jacobian while ours only relies on it for data collection and model selection, which could be avoided if we have a separate sensor for corresponding end-effector positions and if we choose a different model selection criterion. 
In terms of performance, it only achieves around 72$\%$ success rate even without any obstacles as it struggles at joint limits. Contrary to global planners, each action can be executed after each update is computed. Thus, it may appear to be surprisingly slow, while in reality it achieves real time performance.

Overall, it is encouraging to see that LSPP, an intuitive and data-driven formulation, is approaching the performance of established path planning algorithms.

\subsection{Dynamic Feasibility}

\begin{figure*}[th]
\centering
\includegraphics[width=\textwidth]{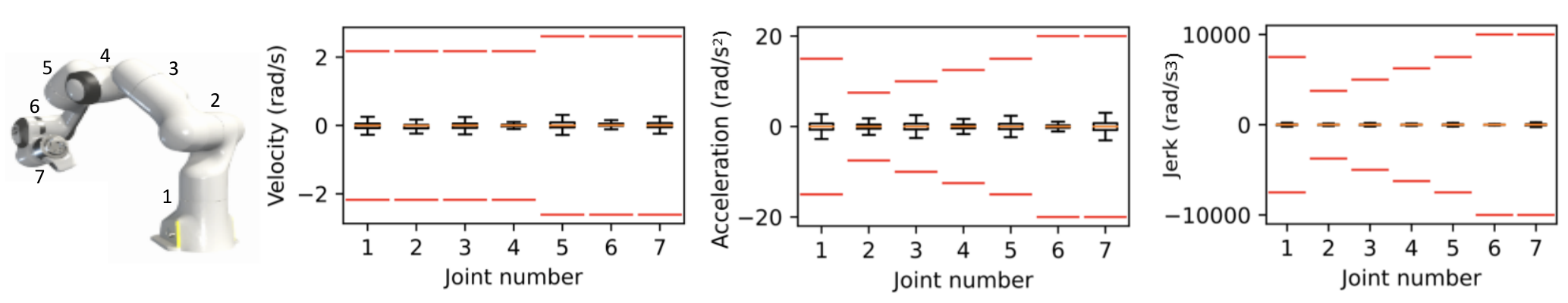}
\caption{Dynamic feasibility of motion plans for Panda arm over 1,000 trajectories. No LSPP motion plans violate the joint limits, indicated by the red segments.
Left: angular velocity.  Middle: angular acceleration. Right: angular jerk.
}
\label{fig:dynamic_feasibility} 
\end{figure*}

To show that the plans are dynamically feasible, we present an analysis in Fig.~\ref{fig:dynamic_feasibility}.
The generated motion plans, when executed with a constant control frequency of $50$Hz, demand a relatively small angular velocity, angular acceleration and angular jerk. These are all well below the maximum joint limits for the Panda arm, shown as red segments in the figure.
This demonstrates that we can generate feasible state space motion plans by decoding from the latent trajectory we obtain from gradient-based optimisation.
Additionally, we can potentially further improve the smoothness by adjusting the learning rate of the Adam optimiser during the optimisation. 

\section{CONCLUSION}\label{sec:discussion}

We present a novel approach to path planning for robot manipulation that learns a structured latent representation of the robot's state space and uses constrained optimisation to produce joint space paths to reach end-effector goals.
Our approach differs significantly from related work in that it performs path planning based on a generative model of robot state, which is trained in a largely task-agnostic manner. In addition to the goal and obstacle losses, we introduce a novel constraint which maximises the likelihood of the latent variable being explored under its learned prior, thereby encouraging the model to stay near the training distribution of robot configurations. In doing so, we bypass the traditional computational challenges encountered by established planning methods while achieving commensurate performance in terms of reaching success, planning time and path length.
Despite the lack of theoretical guarantees, it is a practical mechanism for path planning.
Future directions include algorithmic improvement to handle local minima, generalisation to scenarios with more complex obstacles and dynamic objects, and tasks that involve interaction.

\\



\section*{ACKNOWLEDGMENT}
This work was supported by the UKRI/EPSRC Programme Grant [EP/V000748/1], NIA [EP/S002383/1], the RAIN [EP/R026084/1] and ORCA [EP/R026173/1] Hubs,
the Clarendon Fund and Amazon Web Services as part of the Human-Machine Collaboration Programme. 
The authors also gratefully acknowledge the use of the University of Oxford Advanced Research Computing (ARC) facility in carrying out this work  (\url{http://dx.doi.org/10.5281/zenodo.22558}) and the use of Hartree Centre resources.
We thank Jonathan Gammell for insightful feedback and discussions, and Rowan Border for helping with setting up BIT* and interfacing between OMPL and MoveIt.
We also thank Yizhe Wu for recording real-world experiments, and Jack Collins for proofreading our work.



\bibliographystyle{IEEEtran}
\bibliography{references}

\begin{thebibliography}{10}
\providecommand{\url}[1]{#1}
\csname url@samestyle\endcsname
\providecommand{\newblock}{\relax}
\providecommand{\bibinfo}[2]{#2}
\providecommand{\BIBentrySTDinterwordspacing}{\spaceskip=0pt\relax}
\providecommand{\BIBentryALTinterwordstretchfactor}{4}
\providecommand{\BIBentryALTinterwordspacing}{\spaceskip=\fontdimen2\font plus
\BIBentryALTinterwordstretchfactor\fontdimen3\font minus
  \fontdimen4\font\relax}
\providecommand{\BIBforeignlanguage}[2]{{%
\expandafter\ifx\csname l@#1\endcsname\relax
\typeout{** WARNING: IEEEtran.bst: No hyphenation pattern has been}%
\typeout{** loaded for the language `#1'. Using the pattern for}%
\typeout{** the default language instead.}%
\else
\language=\csname l@#1\endcsname
\fi
#2}}
\providecommand{\BIBdecl}{\relax}
\BIBdecl

\bibitem{lavalle1998rapidly}
S.~M. Lavalle, ``Rapidly-exploring random trees: A new tool for path
  planning,'' Iowa State University, Tech. Rep., 1998.

\bibitem{Kavraki1996prob}
L.~E. {Kavraki}, P.~{Svestka}, J.~. {Latombe}, and M.~H. {Overmars},
  ``Probabilistic roadmaps for path planning in high-dimensional configuration
  spaces,'' \emph{IEEE Transactions on Robotics and Automation}, vol.~12,
  no.~4, pp. 566--580, 1996.

\bibitem{karaman2011sampling}
S.~Karaman and E.~Frazzoli, ``Sampling-based algorithms for optimal motion
  planning,'' \emph{International Journal of Robotics Research}, vol.~30,
  no.~7, pp. 846--894, 2011.

\bibitem{ratliff2009chomp}
N.~Ratliff, M.~Zucker, J.~A. Bagnell, and S.~Srinivasa, ``{CHOMP}: Gradient
  optimization techniques for efficient motion planning,'' in \emph{2009 IEEE
  International Conference on Robotics and Automation}.\hskip 1em plus 0.5em
  minus 0.4em\relax IEEE, 2009, pp. 489--494.

\bibitem{KalakrishnanSTOMP2011}
\BIBentryALTinterwordspacing
M.~Kalakrishnan, S.~Chitta, E.~Theodorou, P.~Pastor, and S.~Schaal, ``{STOMP:
  Stochastic trajectory optimization for motion planning},'' in \emph{2011 IEEE
  International Conference on Robotics and Automation}.\hskip 1em plus 0.5em
  minus 0.4em\relax IEEE, may 2011, pp. 4569--4574. [Online]. Available:
  \url{http://ieeexplore.ieee.org/document/5980280/}
\BIBentrySTDinterwordspacing

\bibitem{ratliff2018riemannian}
N.~D. Ratliff, J.~Issac, D.~Kappler, S.~Birchfield, and D.~Fox, ``Riemannian
  motion policies,'' \emph{arXiv preprint arXiv:1801.02854}, 2018.

\bibitem{Berenson2009}
\BIBentryALTinterwordspacing
D.~Berenson, S.~S. Srinivasa, D.~Ferguson, and J.~J. Kuffner, ``Manipulation
  planning on constraint manifolds,'' in \emph{2009 IEEE International
  Conference on Robotics and Automation}, vol.~5, no.~4.\hskip 1em plus 0.5em
  minus 0.4em\relax IEEE, may 2009, pp. 625--632. [Online]. Available:
  \url{http://ieeexplore.ieee.org/document/5152399/}
\BIBentrySTDinterwordspacing

\bibitem{levine2018learning}
S.~Levine, P.~Pastor, A.~Krizhevsky, J.~Ibarz, and D.~Quillen, ``Learning
  hand-eye coordination for robotic grasping with deep learning and large-scale
  data collection,'' \emph{The International Journal of Robotics Research},
  vol.~37, no. 4-5, pp. 421--436, 2018.

\bibitem{Ichter2019}
\BIBentryALTinterwordspacing
B.~Ichter and M.~Pavone, ``Robot motion planning in learned latent spaces,''
  \emph{IEEE Robotics and Automation Letters}, vol.~4, no.~3, pp. 2407--2414,
  jul 2019. [Online]. Available:
  \url{https://ieeexplore.ieee.org/document/8653875/}
\BIBentrySTDinterwordspacing

\bibitem{Qureshi2019}
A.~H. Qureshi, A.~Simeonov, M.~J. Bency, and M.~C. Yip, ``{Motion planning
  networks},'' \emph{Proceedings - IEEE International Conference on Robotics
  and Automation}, vol. 2019-May, pp. 2118--2124, 2019.

\bibitem{Qureshi2020}
A.~H. Qureshi, J.~Dong, A.~Choe, and M.~C. Yip, ``{Neural manipulation planning
  on constraint manifolds},'' \emph{IEEE Robotics and Automation Letters},
  vol.~5, no.~4, pp. 6089--6096, 2020.

\bibitem{srinivas2018universal}
A.~Srinivas, A.~Jabri, P.~Abbeel, S.~Levine, and C.~Finn, ``Universal planning
  networks: Learning generalizable representations for visuomotor control,'' in
  \emph{International Conference on Machine Learning}.\hskip 1em plus 0.5em
  minus 0.4em\relax PMLR, 2018, pp. 4732--4741.

\bibitem{watter2015embed}
M.~Watter, J.~Springenberg, J.~Boedecker, and M.~Riedmiller, ``Embed to
  control: A locally linear latent dynamics model for control from raw
  images,'' in \emph{Advances in neural information processing systems}, 2015,
  pp. 2746--2754.

\bibitem{banijamali2017robust}
E.~Banijamali, R.~Shu, M.~Ghavamzadeh, H.~Bui, and A.~Ghodsi, ``Robust
  locally-linear controllable embedding,'' \emph{arXiv preprint
  arXiv:1710.05373}, 2017.

\bibitem{hafner2018learning}
D.~Hafner, T.~Lillicrap, I.~Fischer, R.~Villegas, D.~Ha, H.~Lee, and
  J.~Davidson, ``{Learning latent dynamics for planning from pixels},''
  \emph{arXiv preprint arXiv:1811.04551}, 2018.

\bibitem{wu2020imagine}
Y.~Wu, S.~Kasewa, O.~Groth, S.~Salter, L.~Sun, O.~{Parker Jones}, and
  I.~Posner, ``Imagine that! leveraging emergent affordances for 3d tool
  synthesis,'' \emph{arXiv preprint arXiv:1909.13561}, 2020.

\bibitem{mitchell2020steps}
A.~L. Mitchell, M.~Engelcke, O.~{Parker Jones}, D.~Surovik, S.~Gangapurwala,
  O.~Melon, I.~Havoutis, and I.~Posner, ``First steps: Latent-space control
  with semantic constraints for quadruped locomotion,'' \emph{IEEE/RSJ
  International Conference on Intelligent Robots and Systems (IROS)}, pp.
  5343--5350, 2020.

\bibitem{erhan2009visualizing}
D.~Erhan, Y.~Bengio, A.~Courville, and P.~Vincent, ``Visualizing higher-layer
  features of a deep network,'' University of Montreal, Tech. Rep. 1341, Jun.
  2009, also presented at the ICML 2009 Workshop on Learning Feature
  Hierarchies, Montr{\'{e}}al, Canada.

\bibitem{Bialkowski2013}
J.~Bialkowski, S.~Karaman, M.~Otte, and E.~Frazzoli, ``{Efficient collision
  checking in sampling-based motion planning},'' \emph{Springer Tracts in
  Advanced Robotics}, vol.~86, pp. 365--380, 2013.

\bibitem{schulman2014motion}
J.~Schulman, Y.~Duan, J.~Ho, A.~Lee, I.~Awwal, H.~Bradlow, J.~Pan, S.~Patil,
  K.~Goldberg, and P.~Abbeel, ``Motion planning with sequential convex
  optimization and convex collision checking,'' \emph{The International Journal
  of Robotics Research}, vol.~33, no.~9, pp. 1251--1270, 2014.

\bibitem{mukadam2018continuous}
M.~Mukadam, J.~Dong, X.~Yan, F.~Dellaert, and B.~Boots, ``Continuous-time
  gaussian process motion planning via probabilistic inference,'' \emph{The
  International Journal of Robotics Research}, vol.~37, no.~11, pp. 1319--1340,
  2018.

\bibitem{banijamali2018robust}
E.~Banijamali, R.~Shu, H.~Bui, A.~Ghodsi \emph{et~al.}, ``Robust locally-linear
  controllable embedding,'' in \emph{International Conference on Artificial
  Intelligence and Statistics}.\hskip 1em plus 0.5em minus 0.4em\relax PMLR,
  2018, pp. 1751--1759.

\bibitem{karl2016deep}
M.~Karl, M.~Soelch, J.~Bayer, and P.~Van~der Smagt, ``Deep variational bayes
  filters: Unsupervised learning of state space models from raw data,''
  \emph{arXiv preprint arXiv:1605.06432}, 2016.

\bibitem{hafner2019learning}
D.~Hafner, T.~Lillicrap, I.~Fischer, R.~Villegas, D.~Ha, H.~Lee, and
  J.~Davidson, ``Learning latent dynamics for planning from pixels,'' in
  \emph{International Conference on Machine Learning}.\hskip 1em plus 0.5em
  minus 0.4em\relax PMLR, 2019, pp. 2555--2565.

\bibitem{seker2019conditional}
M.~Y. Seker, M.~Imre, J.~H. Piater, and E.~Ugur, ``Conditional neural movement
  primitives.'' in \emph{Robotics: Science and Systems}, vol.~10, 2019.

\bibitem{bocsi2011learning}
B.~{B\'{o}csi}, D.~{Nguyen-Tuong}, L.~{Csat\'{o}}, B.~{Sch\"{o}lkopf}, and
  J.~{Peters}, ``Learning inverse kinematics with structured prediction,'' in
  \emph{2011 IEEE/RSJ International Conference on Intelligent Robots and
  Systems}, 2011, pp. 698--703.

\bibitem{REN2020}
H.~Ren and P.~Ben-Tzvi, ``Learning inverse kinematics and dynamics of a robotic
  manipulator using generative adversarial networks,'' \emph{Robotics and
  Autonomous Systems}, vol. 124, p. 103386, 2020.

\bibitem{whitney1969}
D.~E. {Whitney}, ``Resolved motion rate control of manipulators and human
  prostheses,'' \emph{IEEE Transactions on Man-Machine Systems}, vol.~10,
  no.~2, pp. 47--53, 1969.

\bibitem{goldenberg1985}
A.~{Goldenberg}, B.~{Benhabib}, and R.~{Fenton}, ``A complete generalized
  solution to the inverse kinematics of robots,'' \emph{IEEE Journal on
  Robotics and Automation}, vol.~1, no.~1, pp. 14--20, 1985.

\bibitem{wampler1986}
C.~W. {Wampler}, ``Manipulator inverse kinematic solutions based on vector
  formulations and damped least-squares methods,'' \emph{IEEE Transactions on
  Systems, Man, and Cybernetics}, vol.~16, no.~1, pp. 93--101, 1986.

\bibitem{kingma2013auto}
D.~P. Kingma and M.~Welling, ``Auto-encoding variational bayes,''
  \emph{International Conference on Learning Representations (ICLR)}, 2014.

\bibitem{rezende2014stochastic}
D.~J. Rezende, S.~Mohamed, and D.~Wierstra, ``{Stochastic Backpropagation and
  Approximate Inference in Deep Generative Models},'' \emph{International
  Conference on Machine Learning (ICML)}, 2014.

\bibitem{higgins2017beta}
I.~Higgins, L.~Matthey, A.~Pal, C.~Burgess, X.~Glorot, M.~Botvinick,
  S.~Mohamed, and A.~Lerchner, ``{beta-VAE: Learning Basic Visual Concepts with
  a Constrained Variational Framework},'' \emph{International Conference on
  Learning Representations (ICLR)}, 2017.

\bibitem{rezende2018taming}
D.~J. Rezende and F.~Viola, ``{Taming VAEs},'' \emph{arXiv preprint
  arXiv:1810.00597}, 2018.

\bibitem{kingma2014adam}
D.~P. Kingma and J.~Ba, ``Adam: A method for stochastic optimization,''
  \emph{arXiv preprint arXiv:1412.6980}, 2014.

\bibitem{wilson1927probable}
E.~B. Wilson, ``Probable inference, the law of succession, and statistical
  inference,'' \emph{Journal of the American Statistical Association}, vol.~22,
  no. 158, pp. 209--212, 1927.

\bibitem{khatib1986real}
O.~Khatib, ``Real-time obstacle avoidance for manipulators and mobile robots,''
  in \emph{Autonomous robot vehicles}.\hskip 1em plus 0.5em minus 0.4em\relax
  Springer, 1986, pp. 396--404.

\bibitem{flacco2012depth}
F.~Flacco, T.~Kr{\"o}ger, A.~De~Luca, and O.~Khatib, ``A depth space approach
  to human-robot collision avoidance,'' in \emph{2012 IEEE International
  Conference on Robotics and Automation}.\hskip 1em plus 0.5em minus
  0.4em\relax IEEE, 2012, pp. 338--345.

\bibitem{kuffner2000rrt}
J.~J. Kuffner and S.~M. LaValle, ``{RRT-connect}: An efficient approach to
  single-query path planning,'' in \emph{Proceedings 2000 ICRA. Millennium
  Conference. IEEE International Conference on Robotics and Automation.
  Symposia Proceedings (Cat. No. 00CH37065)}, vol.~2.\hskip 1em plus 0.5em
  minus 0.4em\relax IEEE, 2000, pp. 995--1001.

\bibitem{csucan2009kinodynamic}
I.~A. {\c{S}}ucan and L.~E. Kavraki, ``Kinodynamic motion planning by
  interior-exterior cell exploration,'' in \emph{Algorithmic Foundation of
  Robotics VIII}.\hskip 1em plus 0.5em minus 0.4em\relax Springer, 2009, pp.
  449--464.

\bibitem{bohlin2000path}
R.~Bohlin and L.~E. Kavraki, ``Path planning using lazy prm,'' in
  \emph{Proceedings 2000 ICRA. Millennium Conference. IEEE International
  Conference on Robotics and Automation. Symposia Proceedings (Cat. No.
  00CH37065)}, vol.~1.\hskip 1em plus 0.5em minus 0.4em\relax IEEE, 2000, pp.
  521--528.

\bibitem{janson2015fast}
L.~Janson, E.~Schmerling, A.~Clark, and M.~Pavone, ``Fast marching tree: A fast
  marching sampling-based method for optimal motion planning in many
  dimensions,'' \emph{The International journal of robotics research}, vol.~34,
  no.~7, pp. 883--921, 2015.

\bibitem{gammell2020batch}
J.~D. Gammell, T.~D. Barfoot, and S.~S. Srinivasa, ``Batch informed trees
  (bit*): Informed asymptotically optimal anytime search,'' \emph{The
  International Journal of Robotics Research}, vol.~39, no.~5, pp. 543--567,
  2020.

\bibitem{chitta2012moveit}
S.~Chitta, I.~Sucan, and S.~Cousins, ``{MoveIt!}'' \emph{IEEE Robotics \&
  Automation Magazine}, vol.~19, no.~1, pp. 18--19, 2012.

\bibitem{sucan2012open}
I.~A. Sucan, M.~Moll, and L.~E. Kavraki, ``The open motion planning library,''
  \emph{IEEE Robotics \& Automation Magazine}, vol.~19, no.~4, pp. 72--82,
  2012.

\bibitem{zhao17info}
\BIBentryALTinterwordspacing
S.~Zhao, J.~Song, and S.~Ermon, ``{InfoVAE}: Information maximizing variational
  autoencoders,'' \emph{CoRR}, vol. abs/1706.02262, 2017. [Online]. Available:
  \url{http://arxiv.org/abs/1706.02262}
\BIBentrySTDinterwordspacing

\end{thebibliography}

\clearpage

\appendix

\subsection{Training Details}
\subsubsection{LSPP Hyperparameters}

A grid search is run on the following hyperparameter values. The final values are chosen by sample consistency
as discussed in \cref{sec:sample_cons}.

\begin{table}[ht!]
\begin{center}

\resizebox{\linewidth}{!}{%
\begin{tabular}{l|c}
\toprule
\textsc{Parameter} & \textsc{Value} \\
\hline
\texttt{Number of hidden layers} & 2, \textbf{4}, 8 \\ \hline
\texttt{Units per layer} & 64, 128, 256, 1024, \textbf{2048} \\ \hline
\texttt{Latent dimension} & \textbf{7}, 10, 20, 32, 64 \\ \hline
\multirow{2}{15em}{\texttt{GECO reconstruction target $\tau$}}
 & 0.0001, 0.0002, 0.0004, 0.0006, \textbf{0.0008} \\ 
 & 0.001, 0.0012 \\ \hline
\multirow{2}{12em}{\texttt{Learning rate (GECO)}}
 & 0.001, 0.002, 0.003, 0.004, \textbf{0.005}, 0.006, 0.007 \\
& 0.008, 0.009, 0.01, 0.02, 0.05, 0.1 \\ \hline
\texttt{Learning rate (VAE)} & \textbf{0.0001}, 0.0002, 0.0003, 0.0005, 0.001, 0.01 \\
\bottomrule
\end{tabular}
}

\end{center}
\vskip 0.1in
\caption{Training hyperparameters for grid search. Bold font indicates the values chosen.}
\label{table:traininghyperpara2}
\end{table}

\subsubsection{Choice of Hyperparameters} 
We discuss the effects of some of the hyperparameters on model training and performance.

\textbf{Number of hidden layers and units per layer.}
The number of fully connected hidden layers and the number of units per layer in the neural network affect its capacity, which is important for the VAE in modelling the kinematics relationship and for the obstacle classifier in predicting collision. However, having a network that is too large (e.g. eight hidden layers) is found to lead to instabilities in training and to having diminishing returns in terms of performance. 
Thus, a grid search is conducted on the size of the neural network. For memory efficiency, we choose to perform the grid search on the number of hidden units in powers of two starting from two hidden layers and 64 units per layer. 

\textbf{Latent dimension.}
The number of latent dimensions determines the capacity of the latent space to capture the correlation between the joint angles and the end-effector position. As the expressive power of the decoder is finite, having a small number of latent dimensions is found to create an information bottleneck that prevents the VAE from generating accurate reconstructions. The information preference problem \cite{zhao17info} may also be created if we employ a large number of latent dimensions, which means that many of the latent dimensions may not capture any useful information, which in turn encourages the decoder to ignore the latent encoding. This is also not desirable given our motion planning pipeline is based on latent traversal. Thus, we perform a grid search for the number of latent dimensions, and find that a dimension of seven (i.e. the same as the number of DoFs of the robot) achieves the best performance in terms of our metrics. 

\textbf{GECO reconstruction target.}
The GECO reconstruction target $\tau$ imposed as a constraint via a Lagrange multiplier mechanism in GECO \cite{rezende2018taming} is found to be important for the performance of the VAE model. If the goals are strict (i.e. perfect reconstruction), as the size of the neural network is limited and thus limiting its inference and generation capacity, the MSE term in ELBO overwhelms the KL regulariser due to the Lagrange multiplier, leading to overfitting and a mismatch between the posterior and the prior. On the other hand, if the goals are loose, the reconstruction accuracy becomes poor, leading to worse sample consistency. Thus, we use a grid search on the parameter $\tau$.

\textbf{Learning rate (GECO).}
The learning rate for the GECO Lagrange multiplier determines the responsiveness of the $\lambda$ parameter to the violation of the GECO reconstruction target. The higher the learning rate, the more responsive the $\lambda$ parameter becomes in adjusting the relative weights of the two terms in ELBO in training the VAE. However, a high GECO learning rate leads to instabilities in training. The optimal value is then found through a grid search. 

\subsection{Planning Details}

\subsubsection{Modification of GECO}

The following algorithm is applied to update the individual $\lambda$ parameters using a modification of the GECO algorithm ~\cite{rezende2018taming}. The algorithm is applied to different pairs of loss terms $(-\log p(\bz),\norm{\hat{\bb}, \bb_{\text{target}}}_2)$ and $(\sum_i (-\log (1 - p_\vartheta(\bz, \bo_i))),\norm{\hat{\bb}, \bb_{\text{target}}}_2)$ to compute $\lambda_{\text{prior}}$ and $\lambda_{\text{obs}}$ at each update step.

\begin{algorithm}[h]
    \caption{Update GECO $\lambda$}
    \begin{algorithmic}[1]
        \STATE read current $\lambda^t$\;
        \STATE read loss terms $(l^t_1$, $l^t_2)$\;
        \STATE compute constraint violation $C^t = l^t_1 - \tau_{goal}$\;
        \IF{t=0}
            \STATE initialise moving average $C^0_{ma} = C^0$\;
        \ELSE
            \STATE $C^t_{ma} = \alpha_{ma} C^{t-1}_{ma} + (1-\alpha_{ma}) C^t$\;
        \ENDIF
        \STATE compute update step $\kappa^t = \text{exp}(\alpha_{GECO} C^t_{ma})$ \;
        \STATE update $\lambda^{t+1} = \kappa^t\lambda^t$
    \end{algorithmic}
\end{algorithm}

\subsubsection{Planning Hyperparameters}

A grid search on the planning hyperparameters is run on a validation dataset of obstacle scenarios. The values chosen are given in Table~\ref{table:planninghyperpara}.

\begin{table}[ht!]
\begin{center}
\resizebox{\linewidth}{!}{%
\begin{tabular}{l|c}
\toprule
\textsc{Parameter} & \textsc{Value} \\
\hline
\texttt{Learning rate (AM) $\alpha_{AM}$} & 0.03 \\ \hline
\texttt{Learning rate (GECO)} $\alpha_{GECO}$ & 0.01 \\ \hline
\texttt{Max number of planning steps $T$} & 300 \\ \hline
\texttt{Reaching distance threshold $\gamma$} & 0.01 \\ \hline
\texttt{GECO prior loss target $\tau^{prior}_{goal}$} & 0.4, 0.6, 0.7, 0.8, \textbf{0.9}, 1, 1.2, 1.5, 2 \\ \hline
\texttt{Moving average factor $\alpha^{prior}_{ma}$ for prior loss} & 0.8, 0.9, \textbf{0.95} \\ \hline
\texttt{GECO obstacle loss target $\tau^{obs}_{goal}$} & 0.5, 0.6, \textbf{0.7}, 0.8, 0.9, 1.0, 1.5 \\ \hline
\texttt{Moving average factor $\alpha^{obs}_{ma}$ for obstacle loss} & 0.1, 0.2, 0.3, \textbf{0.4}, 0.6, 0.8, 0.9, 0.95 \\
\bottomrule
\end{tabular}
}

\end{center}
\vskip 0.1in
\caption{Planning hyperparameters. Some hyperparameters are fixed. Bold font indicates the values chosen.}
\label{table:planninghyperpara}
\end{table}

\subsection{Choice of Baselines}
MoveIt~\cite{chitta2012moveit} is the most widely used framework for robot manipulation. The Open Motion Planning Library (OMPL)~\cite{sucan2012open} is a collection of sampling-based motion planning algorithms and is the default planner in MoveIt. In our experiments, seven MoveIt path planning algorithms are chosen for comparison. In the following, we provide a summary (mostly condensed from the OMPL documentation) and the rationale behind our choice.

\textbf{RRTConnect}\cite{kuffner2000rrt} is one of the default planners in MoveIt. It grows two RRTs~\cite{lavalle1998rapidly}, one from the start and one from the goal, and attempts to connect them. It is an improved version of RRT and is probabilistic complete, ensuring a solution will be found if one exists, given sufficient runtime. It is commonly used and best known for its fast convergence, even in high-dimensional spaces.

\textbf{LBKPIECE}\cite{csucan2009kinodynamic} is the other default planner in MoveIt. KPIECE, a sampling-based path planning algorithm designed specifically for planning in high-dimensional spaces, uses a discretisation to guide the exploration of the continuous space. It offers computational advantages by employing projections from the searched space to lower-dimensional Euclidean spaces for estimating exploration coverage. LBKPIECE is a bi-directional variant of KPIECE with lazy collision checking and one level of discretisation. It is also commonly used and known for its planning efficiency.

\textbf{RRT*}\cite{karaman2011sampling} is an asymptotically optimal incremental sampling-based path planning algorithm. It is an optimal variant of RRT and converges to an optimal solution in terms of path length after infinite time. This baseline is insightful in comparing path length.

\textbf{LazyPRM*}\cite{bohlin2000path} is another asymptotically optimal sampling-based planner. The Probabilistic Roadmap Method (PRM) constructs a roadmap and checks whether a path exists in the roadmap between a start and goal state. PRM* gradually increases the number of connection attempts as the roadmap grows in a way that provides convergence to the optimal path. LazyPRM* is a variant of PRM* with lazy state validity checking. This is another useful baseline for path length.

\textbf{FMT*}\cite{janson2015fast} stands for Fast Marching Tree. It is another asymptotically optimal sampling-based planner. The algorithm is specifically aimed at solving complex motion planning problems in high-dimensional configuration spaces, by performing a lazy dynamic programming recursion on a set of probabilistically-drawn samples to grow a tree of paths.

\textbf{BIT*}\cite{gammell2020batch} stands for Batch Informed Trees. It is an anytime asymptotically optimal sampling-based planner that uses heuristics to prioritise expansion towards the goal and high-quality paths. It has been shown to outperform existing sampling-based planning algorithms, e.g. RRT* and FMT*, in terms of computational cost to find equivalent results.

\textbf{CHOMP}\cite{ratliff2009chomp} stands for covariant Hamiltonian optimisation for motion planning. It is a gradient-based trajectory optimisation procedure, and uses two objective functions: an obstacle term that captures obstacle avoidance and a smoothness term that captures the dynamics of the trajectory. It is able to avoid obstacles in most cases, but it can fail if it gets stuck in a local minimum due to a bad initial guess for the trajectory. OMPL can be used to generate collision-free seed trajectories for CHOMP to mitigate this issue. Thus, in our experiments, we use OMPL with the default RRTConnect planner for an initial guess and use CHOMP as a post-processor. CHOMP is efficient, produces smooth paths and is the most commonly used optimisation-based approach.

\subsection{Analysis on Latent Space Representation}

\begin{figure}[!b]
\includegraphics[width=.495\columnwidth]{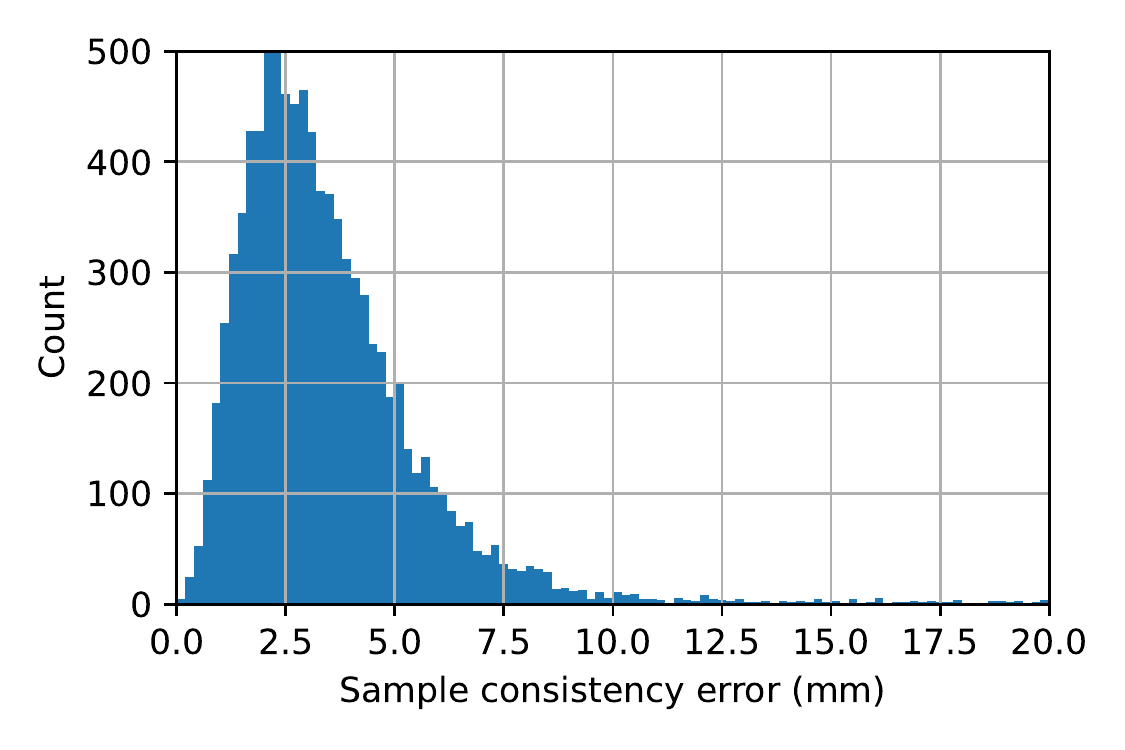}
\includegraphics[width=.495\columnwidth]{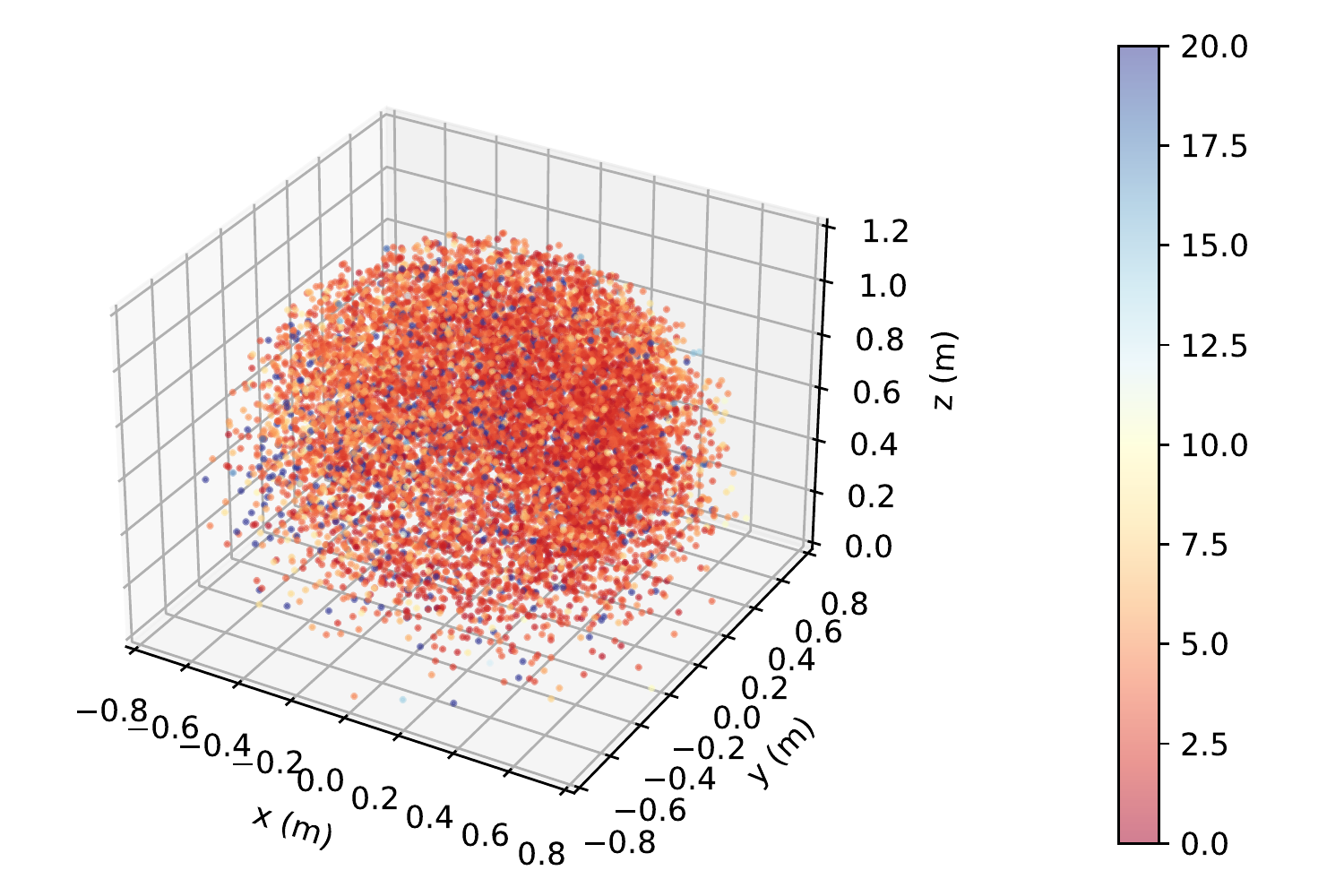}
\caption{Analysis on latent space representation. Left: histogram of sample consistency errors from 10,000 prior samples. Right: 3d scatter of sample consistency errors at different end-effector positions.}
\label{fig:prior_sample_consistency}
\end{figure}

The latent space of our VAE encodes pairs of joint position $\ba$ and end-effector position $\bb$. How accurate is the $\ba$
-to-$\bb$ mapping in the latent space? \Cref{fig:prior_sample_consistency} (left) presents a histogram of sample consistency errors from 10,000 prior samples. The peak is centred at around 2.5mm and over 95\% of the samples fall below 1cm.
What is the completeness of the latent space, i.e. does the latent space cover all the valid joint space or does it miss parts of it? To answer this question, 10,000 samples are drawn from the prior and decoded to $\{\hat{\ba}, \hat{\bb}\}$. In \cref{fig:prior_sample_consistency} (right), we plot the sample consistency errors at different end-effector positions $\hat{\bb}$. We observe that all the valid joint space is well covered and there is no obvious correlation between the sample consistency error and the end-effector position.

\end{document}